\documentclass[letterpaper]{article} % DO NOT CHANGE THIS
\usepackage{aaai24}  % DO NOT CHANGE THIS
\usepackage{times}  % DO NOT CHANGE THIS
\usepackage{helvet}  % DO NOT CHANGE THIS
\usepackage{courier}  % DO NOT CHANGE THIS
\usepackage[hyphens]{url}  % DO NOT CHANGE THIS
\usepackage{graphicx} % DO NOT CHANGE THIS
\usepackage{natbib}  % DO NOT CHANGE THIS AND DO NOT ADD ANY OPTIONS TO IT
\usepackage{caption} % DO NOT CHANGE THIS AND DO NOT ADD ANY OPTIONS TO IT
\usepackage{algorithm}
\usepackage{algorithmic}

\usepackage{amsmath}
\usepackage{subcaption}

\usepackage{newfloat}
\usepackage{listings}
\DeclareCaptionStyle{ruled}{labelfont=normalfont,labelsep=colon,strut=off} % DO NOT CHANGE THIS
\floatstyle{ruled}
\newfloat{listing}{tb}{lst}{}
\floatname{listing}{Listing}
\nocopyright % Technically not allowed, but I do not want to publish this (and retain the rights)

\newcommand\Tstrut{\rule{0pt}{2.6ex}}         % = `top' strut
\newcommand\Bstrut{\rule[-0.9ex]{0pt}{0pt}}   % = `bottom' strut

\graphicspath{{figs/}}

\title{Interpretable Brain-Inspired Representations Improve\\RL Performance on Visual Navigation Tasks
}
\author {
    Moritz Lange\textsuperscript{\rm 1},
    Raphael C. Engelhardt\textsuperscript{\rm 2},
    Wolfgang Konen\textsuperscript{\rm 2},
    Laurenz Wiskott\textsuperscript{\rm 1}
}
\affiliations {
    \textsuperscript{\rm 1}Ruhr-University Bochum, Faculty for Computer Science, Bochum, Germany\\
    \textsuperscript{\rm 2}TH Köln, Cologne Institute for Computer Science, Köln, Germany\\
    moritz.lange@ini.rub.de, raphael.engelhardt@th-koeln.de, wolfgang.konen@th-koeln.de, laurenz.wiskott@rub.de
}

\begin{document}

\maketitle

\begin{abstract}
Visual navigation requires a whole range of capabilities. A crucial one of these is the ability of an agent to determine its own location and heading in an environment. Prior works commonly assume this information as given, or use methods which lack a suitable inductive bias and accumulate error over time. In this work, we show how the method of slow feature analysis (SFA), inspired by neuroscience research, overcomes both limitations by generating interpretable representations of visual data that encode location and heading of an agent. We employ SFA in a modern reinforcement learning context, analyse and compare representations and illustrate where hierarchical SFA can outperform other feature extractors on navigation tasks.
\end{abstract}

\section{Introduction}
Visual navigation is a complex but increasingly relevant task in robotics and in machine learning (ML). Research in this field touches on a wide range of agent capabilities, including the parsing of tasks \cite{wang2021vision}, locating objects to interact with \cite{lyu2022improving}, mapping out the environment \cite{Chaplot2020Learning} and planning \cite{gupta2017cognitive}. A basic capability in navigation, however, is that the agent always has to move around and find a path to its target.

Finding a path to a location, crucially, requires awareness of one's own location and heading. Unsurprisingly, it has been found that an agent's ability of self-localization is important for navigation and especially long-term planning in ML \cite{zhu2021deep}.

In computational neuroscience, slow feature analysis (SFA) \cite{wiskott2002slow} is a method modelled on the human visual system that has long been known for its ability to extract position and head direction from a visual stream. In fact, the representations it generates have been related to place cells and head-direction cells, among others \cite{franzius2007slowness}. In this paper, we utilize a hierarchical model of SFA to extract representations that explicitly encode location and direction of an agent from visual input only. We use these representations to successfully train a reinforcement learning (RL) agent in simple navigation tasks, in order to showcase the potential of these meaningful and explainable representations.

The contributions of this paper are threefold:
\begin{itemize}
    \item We explain how SFA representations significantly differ conceptually from current approaches to localization for visual navigation in RL. Other methods either require integration of information over time or lack a suitable inductive bias for extracting interpretable location and heading information from images. SFA addresses both weaknesses (see Related Work).
    \item We show empirically that SFA representations are not only capable of extracting location and heading, they also make navigation more efficient than other representations which do not contain this information (see Results).
    \item We explain limitations which currently prevent SFA from seamless integration into RL agents, in particular a lack of gradient-based training procedures and the requirements on environment coverage in training data (see Discussion).
\end{itemize}

\section{Related work}
\label{rel_work}
This section explains how previous works on navigation in ML literature have addressed the extraction of location, heading and pose information. It also presents an overview of the relevant prior works on SFA for navigation, which stem from the field of computational neuroscience.

\paragraph{Localization for Navigation}
Representation learning in the context of RL and navigation is often approached through auxiliary tasks \cite{lange2023improving, jaderberg2017reinforcement, ye2021auxiliary2, mirowski2017learning}, often without explicitly considering position, orientation or pose of an agent. The works that do use these features, however, can broadly be split into three categories.

The first, and easy approach is to just assume the agent is provided with ground truth information on its current absolute location and heading \cite{ye2021auxiliary2, ye2021auxiliary}. The second approach can be called location through integration. It is based on the idea that current changes in position and direction can either be inferred or are provided to the agent. These are then integrated over time \cite{mirowski2017learning, Chaplot2020Learning}. The third approach employs neural networks (commonly convolutional neural networks [CNNs] combined with recurrent neural networks [RNNs]) to learn representations from visual input \cite{mousavian2019visual}. These networks do not have any inductive bias towards learning position or heading in particular, although they might be trained in a supervised way directly on this information \cite{wang2017deepvo, datta2021integrating}. Both these papers, additionally, still implicitly integrate changes in location. As a consequence, they share the main weakness of the second approach: accumulation of errors over time. This issue is nicely demonstrated in Figures 4, 6 and 8 of \cite{wang2017deepvo}.

While we see \cite{datta2021integrating} as the most similar approach to ours, their method of how to extract location is fundamentally different. Both theirs and ours work on visual input only, but ours is an unsupervised approach that extracts position and heading based on the idea of extracting slowly varying features. It does not use integration. This makes our approach, to the best of our knowledge, the only one that has a model architecture containing an inductive bias appropriate for agent localization.

\paragraph{Navigation with Slow Feature Analysis}
First introduced by Wiskott \cite{Wiskott-1998b, wiskott2002slow}, the basic SFA method was extended to hierarchical networks, not unsimilar to CNNs, in \cite{Wiskott-2000}. Franzius et al. \cite{franzius2007slowness} show how this hierarchical SFA (hSFA) can be used in combination with independent component analysis (ICA) to extract location and head direction (resembling the neuroscientific concepts of place cells and head direction cells) in a neurologically plausible way from the visual input stream of a simulated animal. Beyond first-person visual input, and potentially also interesting for navigation, the same authors have also used hSFA for object recognition \cite{franzius2011invariant}. Based on this work, \cite{legenstein2010reinforcement} have first applied hSFA to RL: Using hSFA-generated representations, they learn a simple Q-function to make a fish in a tank, seen from above, move to a target. The most recent inspiration for this paper, finally, comes from \cite{schonfeld2013ratlab} who have designed a virtual maze for a virtual rat and use this to extract location and head direction from visual input. Since 2013, the fields of deep learning, reinforcement learning and visual navigation have come a long way. Yet, to the best of our knowledge, there have been no works on visual navigation based on hSFA representations since. A gradient-based approach to SFA has been used in the context of RL \cite{hakenes2019boosting}, but not for navigation.

\section{Learning Slow Features}
\begin{figure*}[t]
\centering
\includegraphics[width=0.8\textwidth]{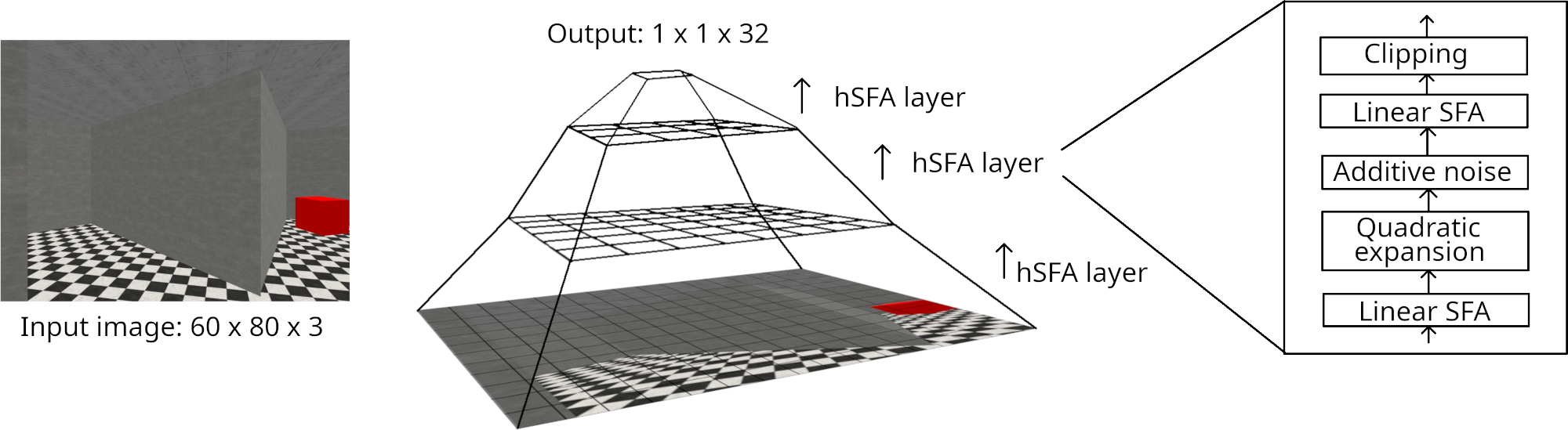} % Reduce the figure size so that it is slightly narrower than the column.
\caption{Illustration of the architecture of a hierarchical slow feature analysis model. The input image is perceived in patches by receptive fields with certain strides. These patches are stacked and passed as batches through an hSFA layer. This happens repeatedly until the last layer produces an output with multiple channels (features), but no width and height.}
\label{fig:hsfa}
\end{figure*}

Slow features in video streams can be extracted with hierarchical slow feature analysis (hSFA), which is based on the underlying method of SFA. We first explain SFA in the linear and then non-linear case, then hSFA.

\paragraph{Slow Feature Analysis}

The idea of slow feature analysis is based on the slowness principle. Invariant or slowly varying features in a signal are usually of more interest than quickly varying features, which are often closer to noise. In a visual stream, for instance, individual pixels will vary very quickly while objects or an agent's position do not. To extract slow features from a signal, SFA solves the following optimization problem:
Given a (commonly multidimensional) signal $x(t)$, find mappings $y_j = g_j(x(t))$ such that
\begin{equation}
    \Delta y_j := \langle \dot{y}_j^2 \rangle_t
\end{equation}
is minimized under the constraints
\begin{align}
    \langle y_j \rangle_t &= 0 \;\;\; \text{(zero mean)}\\
    \langle y_j^2 \rangle_t &= 1 \;\;\; \text{(unit variance)} \\
    \forall i < j: \langle y_iy_j \rangle_t &= 0 \;\;\; \text{(decorrelation and order)}\,.
\end{align}
Here $\langle\rangle_t$ denotes the temporal mean and $\dot{y}$ the temporal derivative of $y$. The extracted signals $y_j$ are the slowest ones which can be created from $x(t)$ given a family of mapping functions $\mathcal{G}$. The constraints guarantee that trivial solutions (a constant signal) are excluded and that output signals are decorrelated and ordered by slowness. For linear SFA, $g_j \in \mathcal{G}$ are chosen to be linear.

In practice, this results in the following algorithm: First, the signal is whitened to obtain zero mean and identity covariance. As an approximation of the temporal derivative, subsequent data points in the time series signal are then subtracted from each other. Lastly, principal component analysis (PCA) is performed on the differentiated time series. The resulting linear components are already decorrelated and ordered by variance. Since their variance now corresponds to the temporal variance in original data, components are ordered by lowest rather than highest variance.

\paragraph{Non-linear SFA}
The family of linear functions is relatively limited in their ability to extract interesting information. Therefore, non-linear expansion is first used on the input signal before performing SFA. This is commonly achieved by quadratic expansion. Even hSFA still uses this expansion as opposed to other non-linearities despite its downside of significantly expanding the dimensionality of data before processing.

\paragraph{Hierarchical SFA}

In order to deal with visual input streams, or videos, hSFA stacks layers of non-linear SFA modules on top of each other (see Figure~\ref{fig:hsfa}). One such layer consists of five components: A linear SFA step first reduces the dimensionality of the data. A quadratic expansion then introduces non-linearity and Gaussian noise is added (during training only) to increase training stability. Finally, another linear SFA extracts the slow features. These features are then clipped, commonly and also in this paper to $[-4, 4]$, to avoid propagation of extreme values. Altogether, this whole hSFA layer is commonly referred to as a step of quadratic SFA.

Each but the top-most layer operates on receptive fields with certain strides, similar to a CNN. Moving a receptive field across the image creates image patches. These patches are flattened and treated as batches to train a hSFA layer, similar to weight sharing in a CNN.

The top-most layer in hSFA is always a quadratic SFA layer that just works on the flattened output of the second-to-top layer. This is comparable to a linear layer at the end of a CNN, it flattens the output and finally allows all parts of the image to have an effect on any dimension of the output.

In contrast to a neural network, the layers of hSFA still, at their core, contain singular value decompositions like PCA. The system is therefore trained layer by layer instead of end-to-end with gradient descent like a usual artificial neural network. Additional control about extracted features can be obtained by using independent component analysis or learning rate adaptation, both of which are described in the Appendix.

\section{Experiments}
In this section, we first describe the RL environments we use to investigate representations and agents. Then we describe how we have trained hSFA, PCA and CNN feature extractors and what agents we test them with.

\subsection{Environments}
We use 3D visual navigation environments of the Miniworld package \cite{MinigridMiniworld23}, which are easy to modify and use. Each environment contains one red cube representing the target. The task is always to reach the target. There are no other objects present.

Observations are  $60 \times 80$ pixel RGB images which show the current front view of the agent in the simulated world. There are three possible actions available: 1) Turn left by $\pi/12$ radians; 2) Turn right by $\pi/12$ radians; 3) Move a small, fixed step forward.

We evaluate performance in terms of episode length $l$ rather than reward $r$. Episode length is a more interpretable measure and contains the same information as the reward, which is calculated as $r=1- 0.2 \frac{l}{l_\text{max}}$. A reward is only made available to the agent once it reaches the box.

Exemplary observations of each environment are shown in Figure~\ref{fig:renderings} in the Appendix. Top-views of their layouts are shown when SFA representations are presented in Figure~\ref{fig:sfa_reps}.

Some of the listed environments are customized, their code is available online \cite{miniworld_repo}.

\paragraph{StarMazeArm}
The target in StarMazeArm is always at the end of the same arm. The initial agent position is a random location in the center room of the maze, its initial heading is random. Maximum episode length is 1500. The optimal policy is to turn until facing the target and then walk forward. In theory this does not require locating the target, as it is always in the same place.

\paragraph{StarMazeRandom}
This environment is identical to StarMazeArm with the exception that the target is placed in a completely random position each episode. The optimal policy is the same as with StarMazeArm. As opposed to StarMazeArm, however, the agent first has to locate the target in each episode before it can know where to walk.

\paragraph{WallGap}
The initial agent position is always in the upper room, the initial target position in the lower room. Initial agent heading is random. Maximum episode length is 300. As opposed to the StarMaze environments, both rooms have the same textures and thus look visually identical apart from one distant skyscraper, which might be visible depending on location and heading. This introduces visual symmetries that make it impossible for many observations to extract position and heading from one image alone. The best policy is to walk straight to the gap between rooms, turn to face the target and walk straight to it.

\paragraph{FourColoredRooms}
The initial agent position and heading are random, as is the target position. Maximum episode length is 250. As opposed to the previous three environments, the wall textures are unique for each wall. Each of the four rooms has a different color, similar to Prince Prospero's rooms in Edgar Allen Poe's The Masque of the Red Death \cite{poe1842masque}. Each wall in a room has a different brightness so that, in contrast to WallGap, there are no visual symmetries despite the symmetry of the layout. The main difficulty is that the number of different rooms makes an exploration strategy necessary to traverse rooms in search of the target.

\subsection{RL Agents}

We train PPO agents with different feature extractors (described below) to solve each navigation task. PPO is a simple, general, state-of-the-art, on-policy, model-free policy optimization algorithm in RL \cite{schulman2017proximal}. Simple here means that it does not involve any navigation capabilities stated in Related work, such as mapping or planning.

We use the implementation of Stable Baselines3 \cite{stable-baselines3} to train five agents with random seeds per setup. Details and hyperparameters can be found in the Appendix. We made all code required to reproduce our experiments and results available on GitHub \cite{github_repo}\footnote{\url{https://github.com/MoritzLange/sfa-for-navigation}}.

In addition to agents trained with feature extractors, we report performance of an agent following random performance and an agent following an optimal policy for comparison. The first quantifies the average episode length achieved by 100 random agents on each environment. The second quantifies average episode length of 10 human trials per environment, exploiting a top-view that includes both agent and target and is not part of the observation. These trials were performed by the authors following the optimal policies described above.

\subsection{Feature Extractors}
We use a hSFA feature extractor, two CNNs and a PCA feature extractor with PPO. They are described below, more details can be found in the Appendix (Table~\ref{tab:parameters}).

\paragraph{hSFA}
The hSFA feature extractor is pre-trained individually for each environment layout, i.e. only once for StarMaze. We use the sklearn-sfa implementation \cite{sklearn-sfa} to extract representations with 32 features. The pretraining is done on 80,000 data points collected by an agent following a random policy. While 80,000 is a high number, such an amount of data is cheap to collect and in this work we focus on demonstrating the capabilities of hSFA rather than exploring the limits of its hyperparameters. Our experience indicates that far fewer points should be sufficient if collecting them were to be expensive. It is however important that they cover a representative sample of combinations of all locations and headings that the agent might later experience.

Training data is collected on empty environments, i.e. we remove the target cube in order to cover all locations and headings, even those that would otherwise be blocked by the target. The hSFA representations are thus not trained on observations with targets, even though we later find indications that the visual cue of a target might still end up being encoded within representations during inference.

Regular resets at maximum episode length of each environment ensure a uniform coverage of the environment, which we found to benefit representations. They do, however, also introduce discontinuities and therefore quick changes in location and heading. We found that these discontinuities do not influence learned representations noticeably if episodes are sufficiently long. While we do not use learning rate adaptation in this work, it could be employed in order to reduce the influence of discontinuities on the representation (see Appendix).

During the training of PPO agents, the pre-trained hSFA feature extractor is used to pre-process the observations fed into the PPO algorithm.

\paragraph{CNNs}
We train two CNN architectures to compare hSFA against. They are prepended to the PPO agent and trained jointly with the agent, i.e. on the RL learning task. The first is NatureCNN \cite{mnih2015human}, the default for processing visual observations in Stable Baselines3. Its purpose is to compare hSFA representations with those that do not have an inductive bias towards encoding location and heading. Additionally, we employ a CustomCNN which mimics the architecture of hSFA. This is to show that the advantage of hSFA stems not from its architecture but from its optimization target.

\paragraph{PCA}
Finally, we train a basic PCA feature extractor. Like hSFA it is also pre-trained, on the same data as hSFA, and then used to pre-process observations when training PPO. We use the scikit-learn implementation \cite{scikit-learn}. The PCA representations consist of 32 features and explain a surprising cumulative amount of variance: 81.9\% for the StarMaze environments, 92.2\% for WallGap and 91.0\% for FourColoredRooms. The purpose of PCA is to show what PPO itself, without any ability to learn complex features, is able to achieve.

\begin{figure*}[t!]
\centering
\begin{subfigure}[t]{\textwidth}
    \centering
    \includegraphics[width=\linewidth]{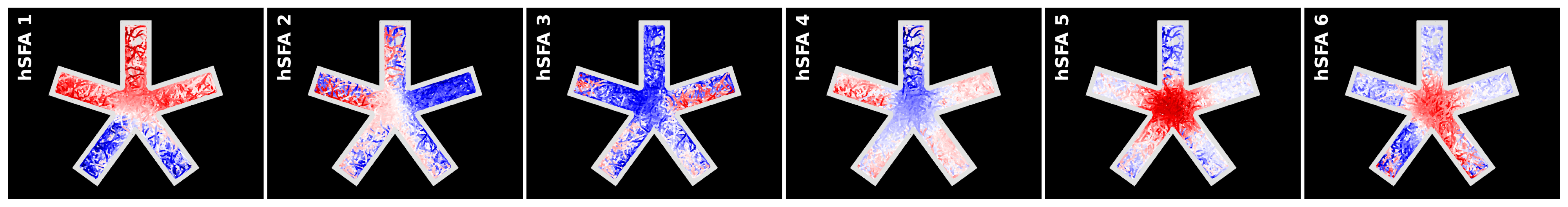} % Reduce the figure size so that it is slightly narrower than the column.
    \caption{First 6 hSFA features for StarMaze}
    \label{fig:sfa_reps_a}
\end{subfigure}%

\begin{subfigure}[t]{\textwidth}
    \centering
    \includegraphics[width=\linewidth]{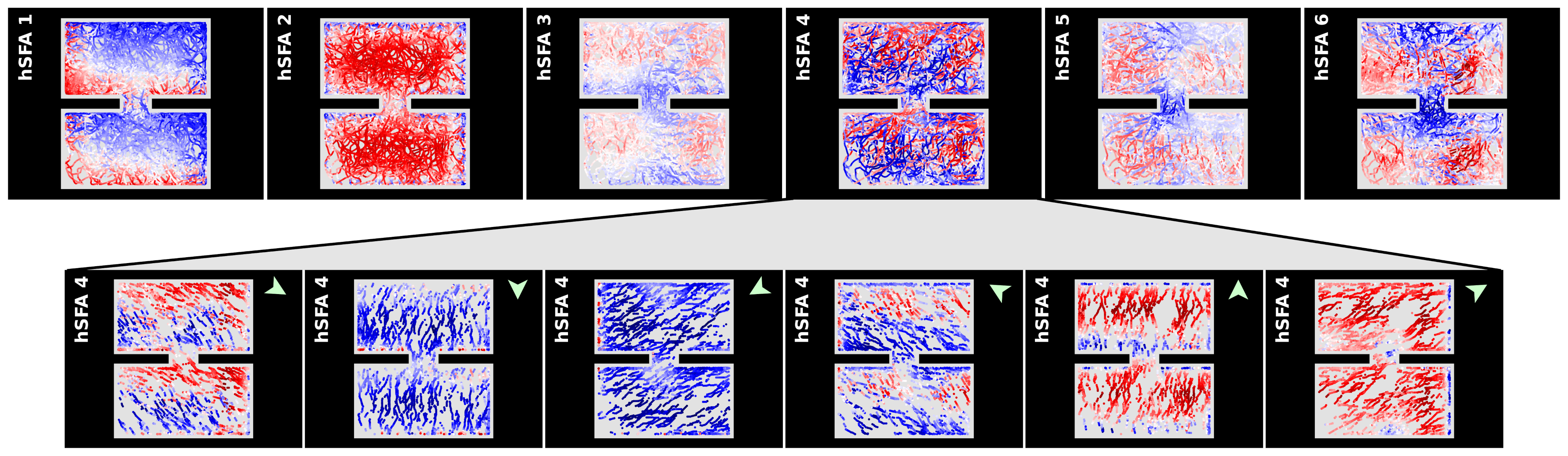} % Reduce the figure size so that it is slightly narrower than the column.
    \caption{First 6 hSFA features for WallGap. Feature 4 is also shown for separate agent headings (green arrow).}
    \label{fig:sfa_reps_b}
\end{subfigure}%

\begin{subfigure}[t]{\textwidth}
    \centering
    \includegraphics[width=\linewidth]{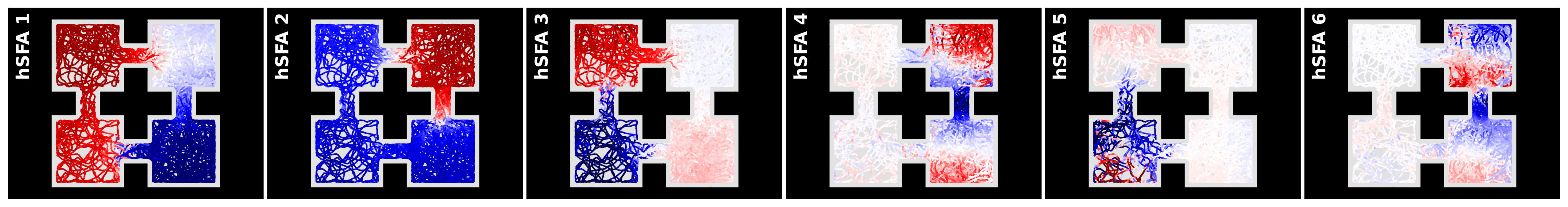} % Reduce the figure size so that it is slightly narrower than the column.
    \caption{First 6 hSFA features for FourColoredRooms}
    \label{fig:sfa_reps_c}
\end{subfigure}%
\caption{Analysis of hSFA representations in different environments (top view). Figures \ref{fig:sfa_reps_a}, \ref{fig:sfa_reps_b}, \ref{fig:sfa_reps_c} show activations of the first 6 hSFA feature dimensions for different positions and orientations in the room. The points are generated by a random agent moving for 80,000 steps without reset. Colors fade from deep red for large positive values into white for zero into deep blue for large negative values. Figure \ref{fig:sfa_reps_b} additionally shows the 4th feature of WallGap for separate agent headings.}
\label{fig:sfa_reps}
\end{figure*}

\section{Results}
We first report the representations obtained with hSFA and then the performance and behaviour of trained agents.

\subsection{Representations}

\begin{figure*}
    \centering
    \includegraphics[width=0.85\textwidth]{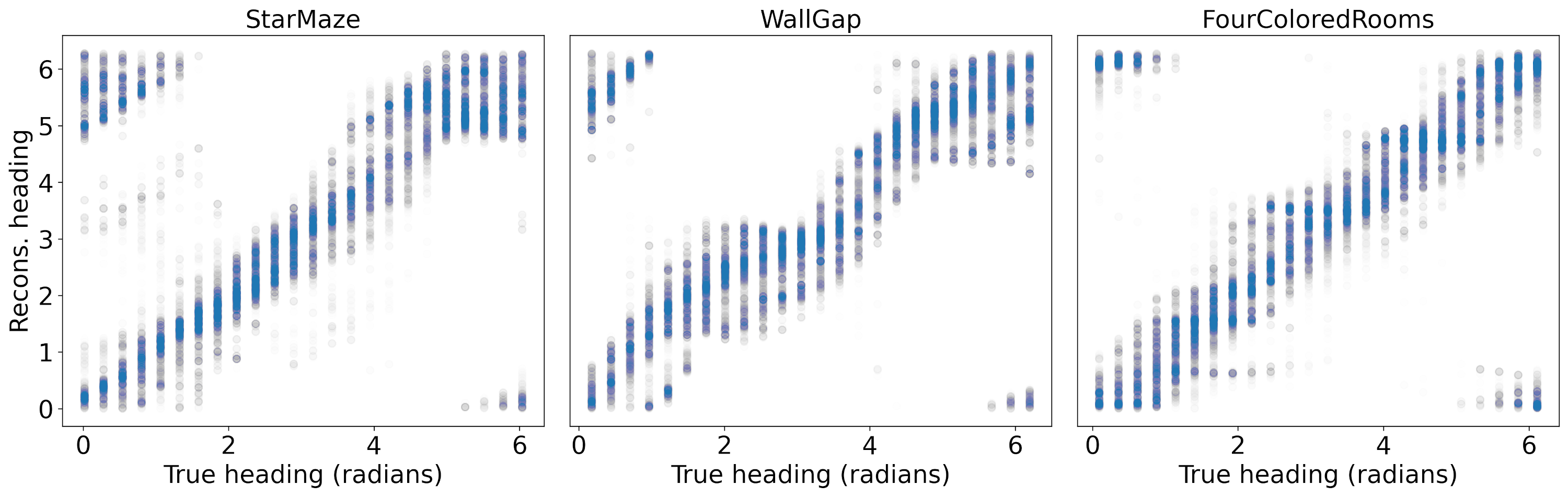}
    \caption{Reconstruction of heading angles. The angle is reconstructed from sine and cosine, which are provided by two linear models trained on all 32 hSFA features. In order to see density, points have a high transparency. The top left and bottom right corners contain points because of the heading's circularity.}
    \label{fig:headings}
\end{figure*}

The representations learned by hSFA are analysed on test sets of 80,000 points, sampled for each environment in the same way as the training data for hSFA was sampled. Information in individual hSFA features is visualized by plotting a top view of the agent's positions and coloring each point by the value of a given feature. Images are shown in Figure~\ref{fig:sfa_reps} for the first~6 (out of~32) hSFA features. Since the train and test set were sampled in the same way, these images additionally provide an intuition on the environment coverage provided by the train set.

In Figures \ref{fig:pca_reps} and \ref{fig:cnn_reps} of the Appendix we also show the representations learned by PCA and NatureCNN for comparison. These were obtained using the same procedure as with hSFA, only the 512 dimensions of the NatureCNN output were additionally passed through a PCA dimensionality reduction to be able to order and display them.

\paragraph{Location}
Figure~\ref{fig:sfa_reps} shows that location information is encoded in the hSFA features. A single feature might activate in different locations, for instance feature~1 of FourColoredRooms has a different relatively constant value inside each room and feature~5 for StarMaze is only strongly positive in the middle room of the maze.

Different components encode different information about location. Earlier components tend to encode global information, later components tend to encode local information. The more components are used, the finer the resulting resolution of location can become.

Additionally we find that components are robust. Even in the StarMaze environment, the maze arms can be confidently differentiated although the only visual information that breaks symmetry is the checkerboard pattern of the floor texture being intersected by walls at different angles (see Figure~\ref{fig:renderings} in the Appendix).

If, however, observations in different positions look exactly identical, this symmetry cannot be resolved by hSFA. This becomes visible for WallGap in Figure~\ref{fig:sfa_reps_b}, where representations are the same in each room. In FourColoredRooms this problem does not arise despite the symmetry of its layout, because each wall has a different color.

Both PCA and NatureCNN are also able to resolve some location information. Their representations however are much more limited in their interpretability. A seemingly meaningful representation of location information is only present in few dimensions and these are considerably noisier than those produced by hSFA.

\paragraph{Heading}
If heading is encoded in a hSFA feature, its visualization becomes a pile of intermingled lines of different colors. The feature takes different values for different headings, and lines arise because even a random agent often walks a couple of subsequent steps into the same direction. This is the case in the fourth hSFA feature for WallGap, shown in the upper row of Figure~\ref{fig:sfa_reps_b}.

In order to investigate the hSFA feature's dependence on heading further, Figure~\ref{fig:sfa_reps_b} in its lower row also shows feature values for different headings. To obtain these, a full circle is divided into 6 angle sections (the arrow indicates the center of each angle section). Each image only shows the agent when its heading is in a given angle section. Feature values are negative when the agent looks south-west, positive when it looks noth-east, and undergo a transition phase between these.

In the case of PCA, this same kind of pattern as for hSFA is also present (see Figure~\ref{fig:pca_reps_b} in the Appendix), although it is less obvious. In the NatureCNN representations, however, Figure~\ref{fig:cnn_reps_b} in the Appendix shows that the noise does not in any way seem to encode heading.

For hSFA, heading generally tends to be encoded in later features (for an explanation, see the ICA and LRA section in the Appendix or \cite{franzius2007slowness}). Heading information is thus not visible in many of the images of Figure~\ref{fig:sfa_reps}. Still, the angle can be reconstructed well from the first 32 hSFA features, as Figure~\ref{fig:headings} shows. This reconstruction, based on linear regression, is accurate to within a few degrees.

To reconstruct heading as in Figure~\ref{fig:headings} and Figures~\ref{fig:pca_headings} and~\ref{fig:cnn_headings} of the Appendix, we learn two linear regressions that map the features to~$\sin(\varphi-\pi)$ and to~$\cos(\varphi-\pi)$, where $\varphi$ is the heading. Its value is then reconstructed from the sine and cosine values. Regressions are trained on the first~40,000 steps of our test data and evaluated on the remaining~40,000 steps. It is necessary to use sine and cosine here because heading is a circular variable with a discontinuity from~$2\pi$ to~$0$. Circular variables (not only angles) are always encoded by their sine and cosine by hSFA, as these do not contain discontinuities and thus vary slowly. The circular nature of the heading is also directly visible in the transition phases for different headings in Figure~\ref{fig:sfa_reps_b}. Despite PCA not relying on slowness, the same reconstruction technique leads to similar results regarding heading for PCA (Figure~\ref{fig:pca_headings}), even though it effectively returns noise for NatureCNN representations (Figure~\ref{fig:cnn_headings}).

\subsection{RL Agents}
\begin{table*}
    \centering
    \begin{tabular}{lr@{\hskip3pt}lr@{\hskip3pt}lr@{\hskip3pt}lr@{\hskip3pt}l}
    \hline
         \Tstrut\Bstrut\textbf{Agent} & \multicolumn{2}{l}{\textbf{StarMazeArm}} & \multicolumn{2}{l}{\textbf{StarMazeRandom}} & \multicolumn{2}{l}{\textbf{WallGap}} & \multicolumn{2}{l}{\textbf{FourColoredRooms}} \\
         %  & StarMazeArm & StarMazeRandom & WallGap & FourColordRooms \\
         \hline
         \Tstrut hSFA & \textbf{69} & \scriptsize{(52, 82)} & \textbf{147} & \scriptsize{(92, 227)} & 277 & \scriptsize{(184, 300)} & 232 & \scriptsize{(225, 239)} \\
         NatureCNN & 415 & \scriptsize{(270, 592)} & 309 & \scriptsize{(226, 430)} &  266 & \scriptsize{(233, 300)} & \textbf{187} & \scriptsize{(178, 201)} \\
         CustomCNN & 652 & \scriptsize{(288, 1487)} & 364 & \scriptsize{(194, 443)} & 300 & \scriptsize{(299, 300)} & 237 & \scriptsize{(211, 250)} \\
         PCA & 773 & \scriptsize{(621, 1069)} & 1005 & \scriptsize{(911, 1099)} & \textbf{179} & \scriptsize{(168, 191)} & 222 & \scriptsize{(212, 233)} \\
         \Bstrut Random & 1134 & \scriptsize{(53, 1500)} & 1073 & \scriptsize{(1, 1500)} & 300 & \scriptsize{(300, 300)} & 231 & \scriptsize{(9, 250)} \\\hline
         \Tstrut\Bstrut Optimal & 36 & & 16 & & 76 & & 53 & \\\hline
    \end{tabular}
    \caption{Average episode lengths achieved by agents with different feature extractors on the different Miniworld environments, at the end of their training. Minimum and maximum of five agents (100 agents for the random policy) are reported in brackets. Best performance is marked in bold. The reported optimal performance is also an average.}
    \label{tab:results}
\end{table*}

The average episode lengths achieved by trained agents are reported in Table~\ref{tab:results}. hSFA agents are much more successful than other agents on the StarMaze environments, but not on WallGap or FourColoredRooms. In the latter two, no agent can be said to perform really well, although the PCA agent outperforms others on WallGap and the NatureCNN agent has a lead in FourColoredRooms.

In the following, we describe observed behaviour for the best out of five agents in each setup. These observations provide a deeper insight than the values in Table~\ref{tab:results}. Since CustomCNN is consistently outperformed by NatureCNN, we only investigate the more successful behaviour of NatureCNN.

\paragraph{StarMazeArm}
The hSFA agent immediately turns in the right direction towards the target and then walks straight to it, even if the target is not immediately visible. Non-optimal performance is explained by the agent sometimes getting stuck at a protruding corner, which is something that regularly happens to all agents across all environments. The PCA agent walks to the target when it is visible, otherwise it wanders into a random direction until it gets stuck in a wall. The NatureCNN agent also walks to the target when visible and walks in circles otherwise.

\paragraph{StarMazeRandom}
The hSFA agent walks in circles around the center room until it sees the target, then walks straight to it. In contrast to this, the NatureCNN agent only spins around itself until it sees the target. If it spawned in a location from which it cannot see the target, it spins until the episode ends. The PCA agent displays the same behaviour as in StarMazeArm.

\paragraph{WallGap}
The hSFA agent sometimes manages to walk directly to the gap connecting both rooms, but more often than not seems confused about the correct direction and ends up walking the wrong direction. If it makes it to the gap, it spins around until it sees the target. If it sees the target it walks towards it. In most cases, it never reaches this last step. The NatureCNN agent walks around randomly until it happens to see the target and then walks straight to it. In many cases the episode ends before it found the target. The PCA agent walks around randomly until it sees the gap. Then it walks straight to the gap. Then it wanders randomly until it sees the cube and walks straight to the cube.

\paragraph{FourColoredRooms}
The hSFA agent walks around almost randomly, often making some distance and covering most of the room it is in. It makes no effort to search for the target in other rooms. If the target becomes visible, the agent does not react to it. Instead the agent seems to rely on hitting the target by walking around. The NatureCNN agent also walks around randomly, however it turns more and covers significantly smaller distances. While it does not actively seem to search for the target, it does walk straight to the target when it becomes visible. The PCA agent walks around randomly and covers as much ground as the hSFA agent, albeit in a seemingly yet more random way. It does however tend to walk to gaps between rooms or to the target when either becomes visible. It does so in a less directed and straightforward way than the NatureCNN agent so that the episode often ends before the target is reached.

\section{Discussion}

We begin our discussion by interpreting our results regarding hSFA representations as well as their use in visual navigation with PPO agents. Then we state current limitations of hSFA, which lead to our the last paragraph outlining potential future work.

\paragraph{Representations}
Our results show that hSFA is concistently able to extract information about both location and heading of the agent, unless there are visual symmetries as in WallGap. We stress again that these symmetries are unlikely outside simple simulated environments.

It is important to note that because the hSFA algorithm directly calculates the solution to a mathematical optimization problem, it is imperative that its outputs are the slowest signals that can be extracted from the input, given the function family $\mathcal{G}$ learnable by its architecture. This stands in contrast to neural networks, where the quality of representations often depends on random seed and initialization \cite{locatello2019challenging}. Furthermore, it implies that if hSFA representations do not encode location and heading, then these are not the slowest signals contained in the visual input stream. Such a thing can happen for various reasons, for instance due to boundary conditions of an environment that result in discontinuities in heading or location.

We conclude that the slowness principle is a valid and powerful inductive bias for extracting location and heading in the visual input stream of an agent. Furthermore, the hSFA algorithm is a suitable architecture to obtain such representations and thus obtains more interpretable representations than those produced by PCA or CNN.

\paragraph{Navigation with hSFA}
The StarMazeArm environment shows that hSFA representations with an explicit and interpretable encoding of location and heading make visual navigation simpler for an RL agent. The agent's movements across all environments except WallGap, in fact, become more purposeful and confident due to the agent's increased awareness of its presence in relation to its surroundings.

In addition, the success on StarMazeRandom illustrates that hSFA representations are able to retain information about the visual scene~--~in this case whether the target cube is visible~--~in addition to location and heading. The fact that the target cube is ignored by the hSFA agent in FourColoredRooms, on the other hand, suggests that extraction of visual cues with current implementations of hSFA has its limit. It has however been shown before that positions of slowly moving objects can be extracted if hSFA is trained on such data \cite{franzius2008invariant}. The question is how slow these features are compared to agent location and heading, and thus how many hSFA features are required to obtain this information.

PCA and CNN are both better at extracting visual cues from an image, as proven by the fact that agents using them walk straight to the target, and sometimes towards gaps, as soon as these become visible. On the other hand, PCA and CNN representations do not or barely encode information about location and heading, as indicated by their comparatively bad performance and inefficient behaviour in StarMazeArm. This is true even for CustomCNN, which supports our claim that the slowness bias rather than the particular architecture of an hSFA feature extractor is responsible for learning location and heading.

In general, it can also be said that PPO is able to solve simple navigation problems when given sufficient representations. That it requires reasonable representations becomes clear by the comparatively bad performance on simple StarMaze environments with PCA representations, which only compress images into the same dimensionality provided by hSFA. For more complex tasks and environments, such as FourColoredRooms however, navigation with a simple RL agent achieves its limits even with meaningful representations. Here, additional capabilities become important, such as planning or mapping.

\paragraph{Limitations}

The main limitation of hSFA is that it always only considers the current observation without any context. It shares this limitation with all other approaches we consider in this work. For successful navigation even in complex settings, it will be important to include an awareness of history or context. All investigated approaches calculate representations from individual observations and PPO only learns a direct mapping from state to action. The ability of executing a plan and keeping some sort of memory, as is the case in planning approaches, will help in addressing this. Future approaches built on hSFA might address this on a representation level to extract location and heading despite symmetries, or on an agent level to interpret representations in context.

Another disadvantage of hSFA, when compared to neural networks, is that it has to be trained layer-wise. This training is slow because it has not been as optimized as gradient descent has, for instance in PyTorch. Additionally, the quadratic feature expansion slows hSFA down. Beyond training of hSFA, this means that training an RL algorithm with observations that first get processed by a trained hSFA feature extractor is about half as fast as training an RL algorithm with prepended CNN feature extractor. A gradient-based version of SFA, which could potentially address both issues, has recently been proposed \cite{schuler2019gradient}.

Finally, the data used to train hSFA has to provide a reasonable coverage of positions and headings in an environment. While this is usually easy to obtain by moving around in a random manner for a short while, it would be nice to learn location and heading online, during exploration of a new area. This is currently not possible.

\paragraph{Future work}
This paper has shown how hSFA can be used to extract location and heading information for visual navigation purposes. The next step will be to combine this functionality with other capabilities an agent requires for complex navigation, in order to see how a state-of-the-art agent with hSFA localization compares to one without. For such an investigation, an hSFA feature extractor can be combined with other feature extractors used e.g. for scene understanding, with language models that decode tasks, and with other modules. It would also be interesting to combine hSFA features with planning approaches for navigation.

Another interesting research direction will be how to overcome the stated limitations of hSFA itself. In particular its requirement to be trained separately rather than end-to-end and the requirement to have seen a representative subsample of the environment during training. With gradient-based SFA already being available, we predict that further research will enable hSFA to be trained online and end-to-end as part of an agent that explores new environments, similar to how animals learn an internal model of new places as they go there.

As a first step in this direction, it will also be interesting to investigate the transfer of representations, i.e. how well a hSFA feature extractor trained in one environment functions in another one.

Lastly, hSFA representations have the potential to make RL agent decisions more explainable. We have focused in this work on the representations and shown how they are significantly more interpretable than those of the other presented approaches that process image data. We have used a classical black-box RL algorithm for the navigation problem which we do not further try to explain here. For future work, however, our findings open up multiple avenues for explainable agents. Most importantly, meaningful representations usually require simpler agents, as these do not have to perform any kind of internal representation learning themselves anymore, focusing purely on decision-making. Additionally, even in the absence of simple agents, the input-output relations of more complex models like PPO become easier to interpret with existing explainability methods when the input is easy to interpret.

\section*{Acknowledgements}
This research was supported by the research training group ``Dataninja'' (Trustworthy AI for Seamless Problem Solving: Next Generation Intelligence Joins Robust Data Analysis) funded by the German federal state of North Rhine-Westphalia.

\bibliography{paper}

\section*{Appendix}

\subsection{Agent Training}
To train our agents, we use the PPO implementation provided by Stable Baselines3 \cite{stable-baselines3}. Parameters which are different from default are reported in Table~\ref{tab:parameters_ppo}. In addition to the reported parameters, we use the CnnPolicy for the NatureCNN feature extractor and the MlpPolicy for the other ones. We train for 1,000,000 steps on the StarMazeArm environment and for 2,000,000 steps on the remaining environments.

\begin{figure*}[t]
\centering
\includegraphics[width=1.0\textwidth]{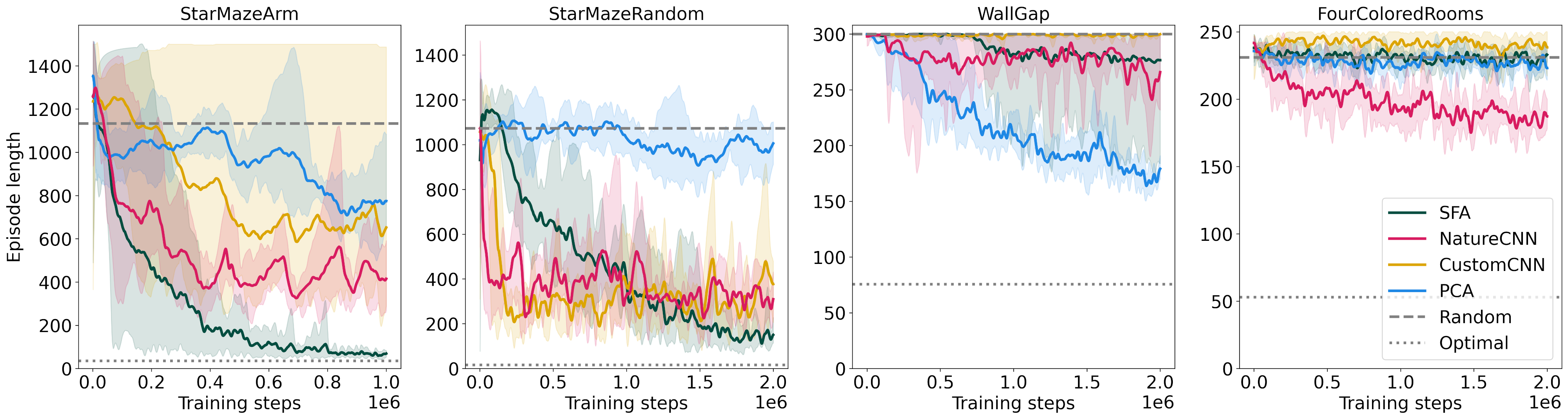} % Reduce the figure size so that it is slightly narrower than the column.
\caption{Performance of agents with various feature extractors on the different Miniworld environments. Shaded areas indicate the minimum and maximum of five agents trained with different random seeds. Curves have been smoothed slightly for clearer presentation.}
\label{fig:results_agents}
\end{figure*}

The layer specifications of the hSFA, NatureCNN and CustomCNN feature extractors are listed in Table \ref{tab:parameters}. hSFA and PCA are pre-trained, and when training the PPO agent they are only used to transform observations into a 32-dimensional feature space. NatureCNN and CustomCNN are trained end-to-end together with the PPO agent on the RL task. Representations by hSFA, PCA and CustomCNN are fed through two fully connected layers of a neural network, as is common for PPO in Stable Baselines3. These layers are identical for all three, trained together with the PPO agent on the RL task, and are listed in Table~\ref{tab:parameters}. NatureCNN, on the other hand, is already an internal feature extractor of Stable Baselines3 and its representations are directly used for policy learning. In addition to the results in Table~\ref{tab:results} of the main paper, we show the training curves for all agents on all environments in Figure \ref{fig:results_agents}.

\begin{table}[]
    \centering
    \begin{tabular}{ll}
    \hline
         \Tstrut\Bstrut \textbf{Parameter} & \textbf{Value} \\\hline
         \Tstrut n\_steps & 128 \\
         learning\_rate & 0.00025 \\
         ent\_coef & 0.01 \\
         clip\_range & 0.1 \\
         \Bstrut batch\_size & 128 \\\hline
    \end{tabular}
    \caption{Parameters used with the PPO model of Stable Baselines3. Only parameters that were not left at their default setting are listed.}
    \label{tab:parameters_ppo}
\end{table}

\begin{table*}
    \centering
    \begin{tabular}{llcccc}
    \hline
         \Tstrut\Bstrut \textbf{Layer} & \textbf{Type} & \textbf{Receptive field} & \textbf{Stride} & \textbf{Exp. deg.} & \textbf{\# Channels out}  \\
         \hline
         \Tstrut hSFA layer 1 & Quadratic SFA & (10, 10) & (5, 5) & 2 & 32  \\
         hSFA layer 2 & Quadratic SFA & (3, 3) & (2, 3) & 2 & 32  \\
         hSFA layer 3 & Quadratic SFA (fully connected) & -- & -- & 2 & 32 \\
         hSFA MLP 1 & Fully connected & -- & -- & -- & 64\\
         \Bstrut hSFA MLP 2 & Fully connected & -- & -- & -- & 64 \\
         \hline
         \Tstrut PCA MLP 1 & Fully connected & -- & -- & -- & 64\\
         \Bstrut PCA MLP 2 & Fully connected & -- & -- & -- & 64\\\hline
         \Tstrut NatureCNN layer 1 & Convolution & (8, 8) & (4, 4) & -- & 32 \\
         NatureCNN layer 2 & Convolution & (4, 4) & (2, 2) & -- & 64  \\
         NatureCNN layer 3 & Convolution & (3, 3) & (1, 1) & -- & 64  \\
         \Bstrut NatureCNN layer 4 & Fully connected & -- & -- & -- & 512  \\
         \hline
         \Tstrut CustomCNN layer 1 & Convolution & (10, 10) & (5, 5) & -- & 32  \\
         CustomCNN layer 2 & Convolution & (3, 3) & (2, 2)& -- & 32  \\
         CustomCNN layer 3 & Convolution & (1, 1) & (1, 1) & -- &  32  \\
         CustomCNN MLP 1 & Fully connected & -- & -- & -- & 64  \\
         \Bstrut CustomCNN MLP 2 & Fully connected & -- & -- & -- & 64  \\
         \hline
    \end{tabular}
    \caption{Parameters used for the hSFA and CNN networks. The MLP layers used with hSFA, PCA and CustomCNN are those introduced by the MlpPolicy in Stable Baselines3. They are automatically appended to the hSFA and CustomCNN feature extractors. In the sklearn-sfa package \cite{sklearn-sfa}, the hSFA layer 3 does not have to be specified. Exp. deg. refers to the degree of expansion, a parameter used in hSFA layers.}
    \label{tab:parameters}
\end{table*}

\begin{figure*}[t!]
    \centering
    \includegraphics[width=0.8\textwidth]{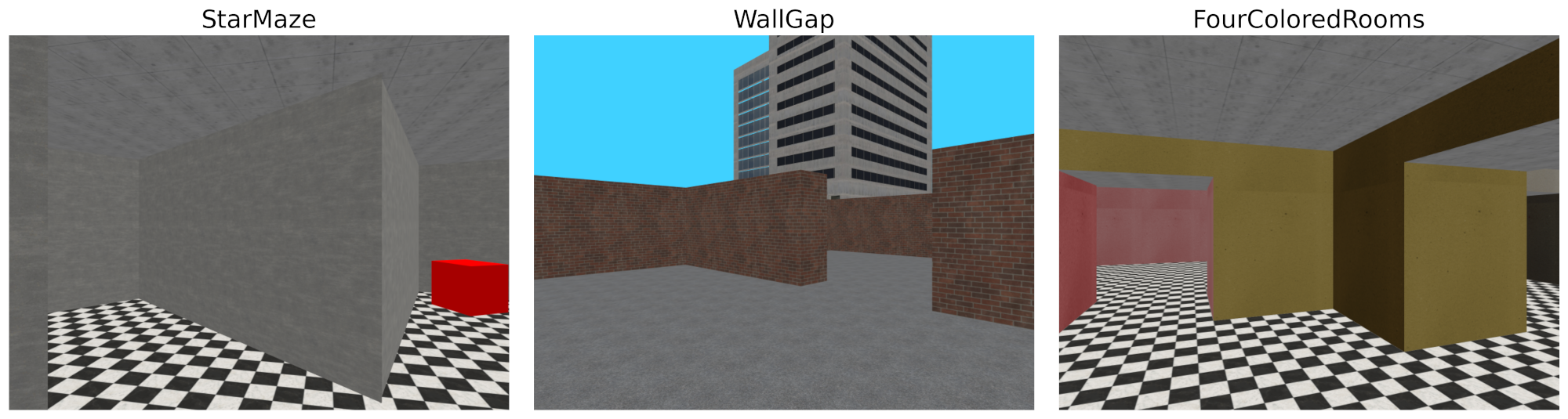}
    \caption{Exemplary observations rendered from the different environments. In this observation of StarMaze, a red target cube is visible. There is no shade from illumination, so the different wall texture colors in FourColoredRooms are in fact textures of different colors.}
    \label{fig:renderings}
\end{figure*}

\subsection{ICA and LRA}
In addition to the base hSFA algorithm, various extensions have been proposed \cite{Escalante-B.Wiskott-2012a}. Two of these can be used to affect the kind of representations extracted from the input. Both have been used to extract of location and heading.

The first is a final layer of sparse coding achieved through independent component analysis (ICA), which is attached to the top of the hSFA model by \cite{franzius2007slowness, schonfeld2013ratlab}. The reason for its use is that only this step of sparse coding transforms location information into neuro-plausible place fields. Additionally, \cite{schonfeld2013ratlab} claim that the use of ICA was required to disentangle head directions. We find, however, that we can obtain head direction and location information without ICA in this work.

The second is learning rate adaptation (LRA) \cite{franzius2007slowness}. By weighing data points, their influence on the SFA results can be controlled. In practice, this is achieved by including weights for the differentiated time series when calculating the covariance matrix that is used for singular value decomposition within SFA \cite{price1972extension}. To calculate weights for the differences between two points, LRA requires an aggregation method. Geometric mean is a good choice, as the arithmetic mean has a tendency to smooth weights out. LRA should be used if there are sudden, fast and unnatural changes in a signal. Two prominent examples are a suddenly mirrored or reflected heading when bouncing into the wall of a simulated environment or an interval of missing data in a time series. These discontinuities would artificially make signals, such as the heading in this example, change faster than they actually do. Such large differences then effectively act similar to how an outlier would in PCA. They strongly affect the whole representation, unless they are mitigated by a small weight. In some cases, it might be practical to apply LRA to certain movements, in particular fast rotations, since rotation of an agent typically changes faster than location when they are normalized by $2\pi$ and size of the environment, respectively. This is proven in \cite{franzius2007slowness}.

LRA is difficult to use if there are only few discrete actions, as is the case in this work. Larger weights for rotations and smaller weights for moving ahead average each other out for almost all differences of data points collected by a random policy as one action is very often followed by a different one. Additionally, we find that we do not need to use LRA under the conditions examined in this work to obtain good representations.

\subsection{Representations Learned by PCA and CNN}
\label{sec:pca_cnn_reps}
\begin{figure*}[t!]
\centering
\begin{subfigure}[t]{\textwidth}
    \centering
    \includegraphics[width=\linewidth]{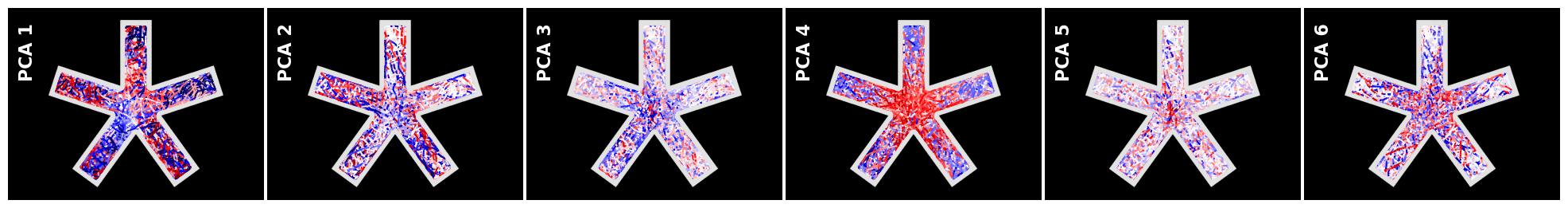} % Reduce the figure size so that it is slightly narrower than the column.
    \caption{First 6 PCA features for StarMaze}
    \label{fig:pca_reps_a}
\end{subfigure}%

\begin{subfigure}[t]{\textwidth}
    \centering
    \includegraphics[width=\linewidth]{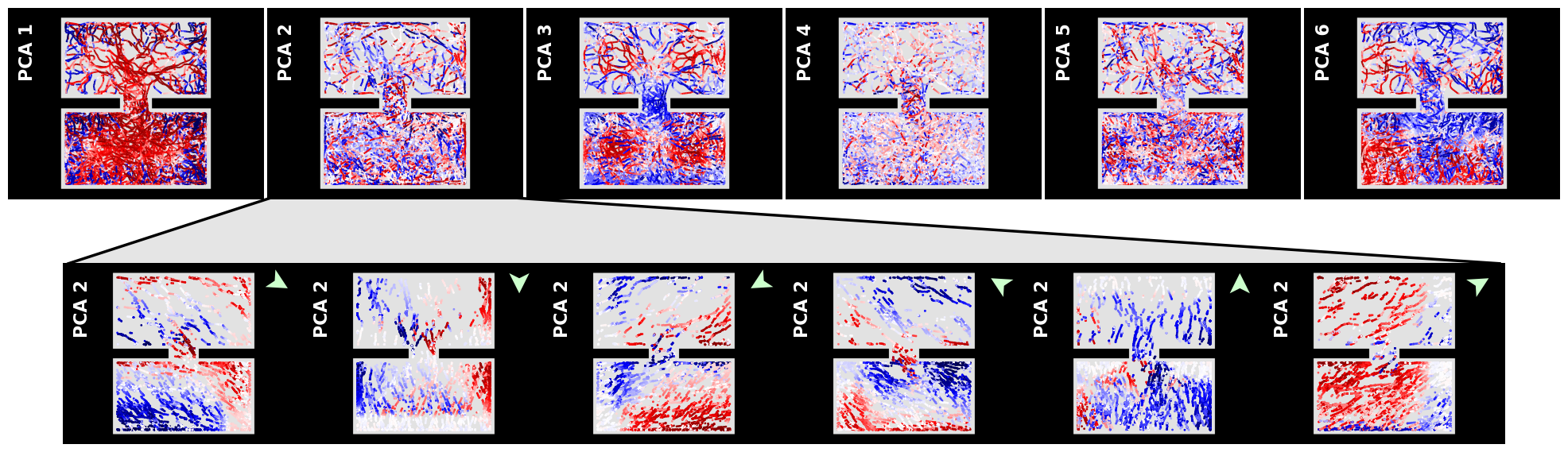} % Reduce the figure size so that it is slightly narrower than the column.
    \caption{First 6 PCA features for WallGap. Feature 2 is also shown for separate agent headings (green arrow).}
    \label{fig:pca_reps_b}
\end{subfigure}%

\begin{subfigure}[t]{\textwidth}
    \centering
    \includegraphics[width=\linewidth]{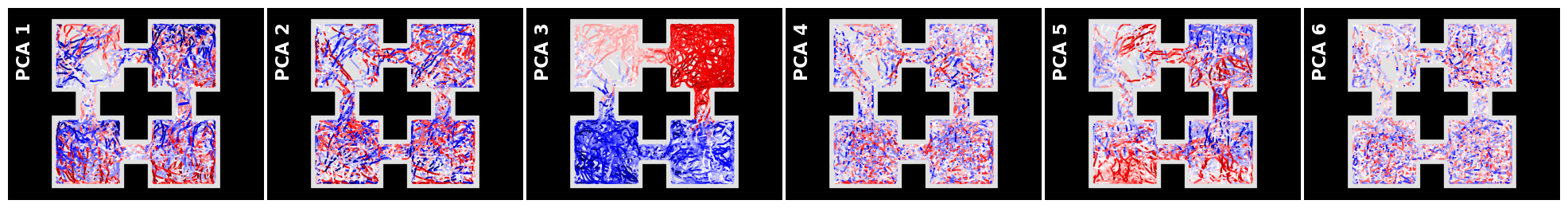} % Reduce the figure size so that it is slightly narrower than the column.
    \caption{First 6 PCA features for FourColoredRooms}
    \label{fig:pca_reps_c}
\end{subfigure}%
\caption{Analysis of PCA representations in different environments (top view). Figures \ref{fig:pca_reps_a}, \ref{fig:pca_reps_b}, \ref{fig:pca_reps_c} show activations of the first 6 PCA feature dimensions for different positions and orientations in the room. The points are generated by a random agent moving for 80,000 steps without reset. Colors fade from deep red for large positive values into white for zero into deep blue for large negative values. Figure \ref{fig:pca_reps_b} additionally shows the 2nd feature of WallGap for separate agent headings.}
\label{fig:pca_reps}
\end{figure*}

\begin{figure*}
    \centering
    \includegraphics[width=0.85\textwidth]{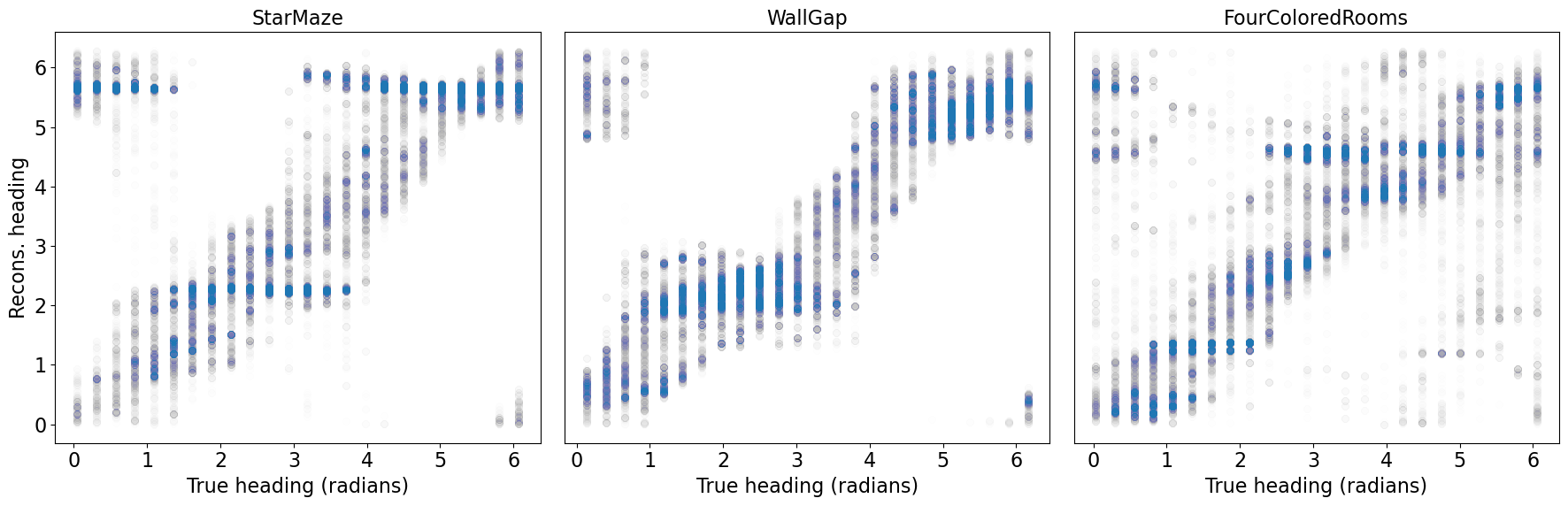}
    \caption{Reconstruction of heading angles for PCA. The angle is reconstructed from sine and cosine, which are provided by two linear models trained on all 32 PCA features. In order to see density, points have a high transparency. The top left and bottom right corners contain points because of the heading's circularity.}
    \label{fig:pca_headings}
\end{figure*}

\begin{figure*}[t!]
\centering
\begin{subfigure}[t]{\textwidth}
    \centering
    \includegraphics[width=\linewidth]{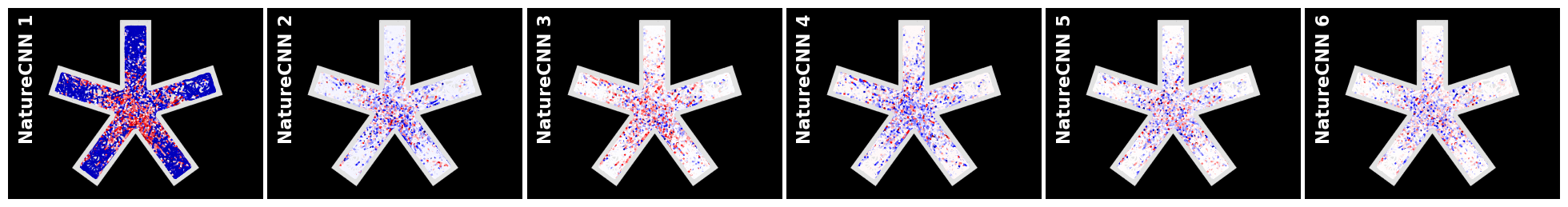} % Reduce the figure size so that it is slightly narrower than the column.
    \caption{First 6 NatureCNN features for StarMaze}
    \label{fig:cnn_reps_a}
\end{subfigure}%

\begin{subfigure}[t]{\textwidth}
    \centering
    \includegraphics[width=\linewidth]{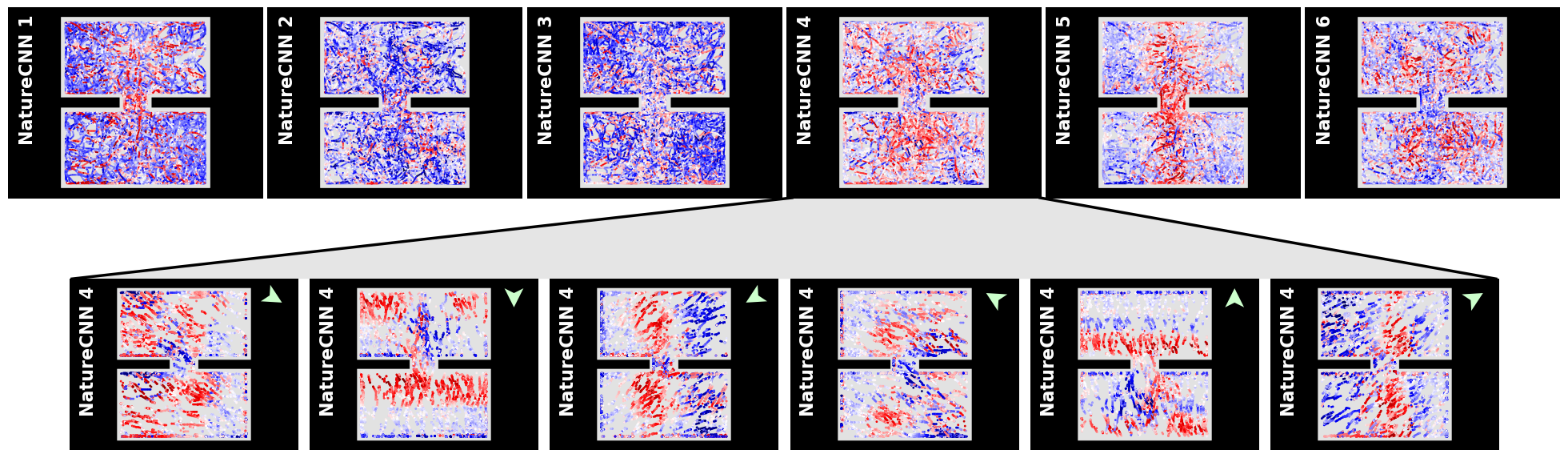} % Reduce the figure size so that it is slightly narrower than the column.
    \caption{First 6 NatureCNN features for WallGap. Feature 4 is also shown for separate agent headings (green arrow).}
    \label{fig:cnn_reps_b}
\end{subfigure}%

\begin{subfigure}[t]{\textwidth}
    \centering
    \includegraphics[width=\linewidth]{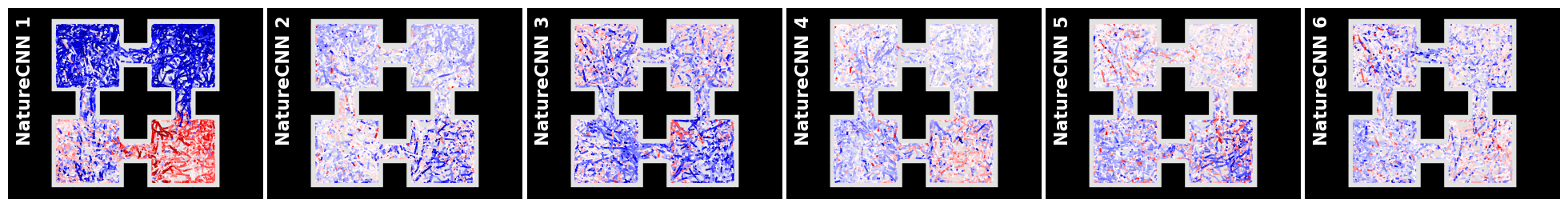} % Reduce the figure size so that it is slightly narrower than the column.
    \caption{First 6 NatureCNN features for FourColoredRooms}
    \label{fig:cnn_reps_c}
\end{subfigure}%
\caption{Analysis of NatureCNN representations in different environments (top view). The raw unordered 512 NatureCNN features are additionally passed through a PCA dimensionality reduction to obtain more meaningful and ordered visualizations, so these Figures do not show the raw representations returned by NatureCNN. Figures \ref{fig:cnn_reps_a}, \ref{fig:cnn_reps_b}, \ref{fig:cnn_reps_c} show activations of the first 6 PCA components for different positions and orientations in the room. The points are generated by a random agent moving for 80,000 steps without reset. Colors fade from deep red for large positive values into white for zero into deep blue for large negative values. Figure \ref{fig:cnn_reps_b} additionally shows the 4th feature of WallGap for separate agent headings.}
\label{fig:cnn_reps}
\end{figure*}

\begin{figure*}
    \centering
    \includegraphics[width=0.85\textwidth]{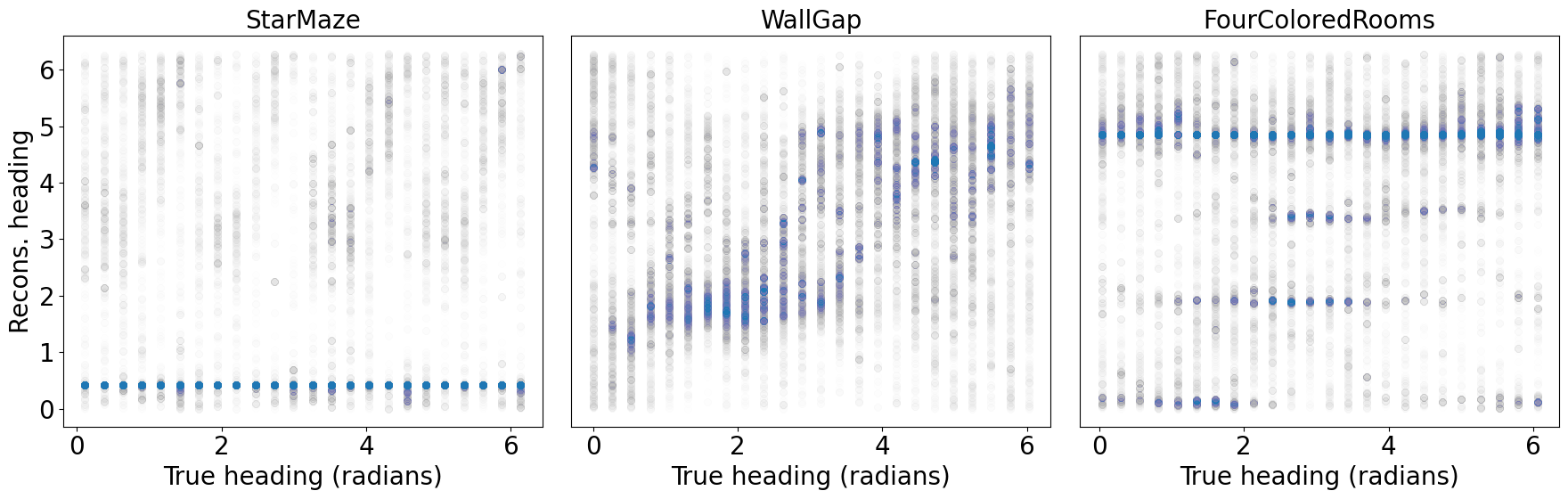}
    \caption{Reconstruction of heading angles for NatureCNN. The angle is reconstructed from sine and cosine, which are provided by two linear models trained on the first 32 dimensions of the output of a PCA dimensionality reduction of the 512-dimensional NatureCNN representations. In order to see density, points have a high transparency.}
    \label{fig:cnn_headings}
\end{figure*}

Here, we present the representations learned by the PCA feature extractor (Figures \ref{fig:pca_reps} for location and \ref{fig:pca_headings} for heading) and the NatureCNN feature extractor (Figures \ref{fig:cnn_reps} for location and \ref{fig:cnn_headings} for heading). As in the Discussion in the main paper, we omit CustomCNN here due to its inferior performance. Since the dimensionality of the NatureCNN representations is 512, and these have no natural order, we apply a PCA dimensionality reduction to make information contained in the representations more concise and also ordered. For StarMaze, WallGap and FourColoredRooms the first six PCA components capture a cumulative variance ratio of 68\%, 89\% and 75\%, respectively. Since the NatureCNN representations are trained together with the RL algorithm instead of independently, the representations in Figure \ref{fig:cnn_reps} are those obtained by the best performing RL agent on each environment.

\end{document}